\documentclass[a4paper, 10pt, conference]{ieeeconf}      
\usepackage[T1]{fontenc}

\IEEEoverridecommandlockouts                              

\usepackage{graphics} 

\usepackage[dvipdfmx]{graphicx}
\usepackage{cite}

\title{\LARGE \bf
A Pin-Array Structure for Gripping and Shape Recognition \\ of Convex and Concave Terrain Profile
}

\author{Takuya Kato$^{1}$, Kentaro Uno$^{1}$ and Kazuya Yoshida$^{1}$
\thanks{$^{1}$The authors are with the Department of Aerospace Engineering, Graduate School of Engineering, Tohoku University, Sendai 9808579, Japan.}
\thanks{{\tt\small kato.takuya.s8@dc.tohoku.ac.jp,}}%
\thanks{{\tt\small \{unoken, yoshida.astro\}@tohoku.ac.jp}}%
}

\begin{document}

\maketitle
\thispagestyle{empty}
\pagestyle{empty}

\begin{abstract}
This paper presents a gripper capable of grasping and recognizing terrain shapes for mobile robots in extreme environments. On rough terrains such as cliffs and cave walls, multi-limbed climbing robots with grippers are effective. However, such robots may fall over due to
misgrasping the surface or get stuck due to the loss of graspable points in unknown natural environments. To overcome such issues, we need a gripper capability of adaptive grasping to irregular terrain, and not only for grasping but also for measuring the shape of the terrain
surface accurately. We have developed a gripper that can grasp both convex and concave terrains and measure the terrain shape at the same time, by introducing a pin-array structure. We show the mechanism of the gripper and evaluate its grasping and terrain recognition performance
using a prototype. We demonstrate the proposed pin-array design works well for 3D terrain mapping, as well as adaptive grasping for irregular terrain.
\end{abstract}
\section{INTRODUCTION}
Mobile robots locomote effectively by utilizing the interaction with an environment intelligently\cite{brooks,pfeifer}. Particularly in unstructured and extreme environments, robots need to acquire environmental information accurately and efficiently to feedback them to the controller. Wheels are widely used for mobility on relatively smooth and non-steep terrain because of the simplicity of the control and mechanism, and energy efficiency. However, it is too tough for the wheeled robots to traverse rigorous natural terrain such as cliffs, ocean floors, and asteroid surfaces. In such natural environments, unevenness of the terrain and the gravitational effect make it difficult to maintain contact with the terrain surface and the robot pose. The addition of adhesivity to grasp the terrain surface helps to maintain the reaction force from the terrain surface and to support the robot body stably. 
Meanwhile, the legged robot's high traversability in natural uneven and steep terrain is widely demonstrated and showing the robustness \cite{wild_anymal}. Therefore, various legged robots with grippers installed at the feet are considered to be the effective solution to realize the traversing and climbing capability of the mobile robots\cite{yoshida2002,scaler,uno_humanoids}.\\
\indent
For stable and efficient locomotion of these climbing robots, it is essential to gain and utilize the appropriate reaction force through the interaction between the gripers and the terrain surface. In this research field, Shirai \it{et al.} \rm have proposed contact-rich grasping/locomotion planning with the terrain surface\cite{shirai_iros}. Another quadrupedal climbing robot: HubRobo\cite{uno_humanoids} developed by our research group is capable of sustaining the foot contact and robot pose by installing the passive spine gripper mechanisms\cite{nagaoka_ral}. \\
\begin{figure}[t]
\renewcommand{\baselinestretch}{0.8}
\vspace{-3mm}
  \centering
  \includegraphics[width=0.95\linewidth]{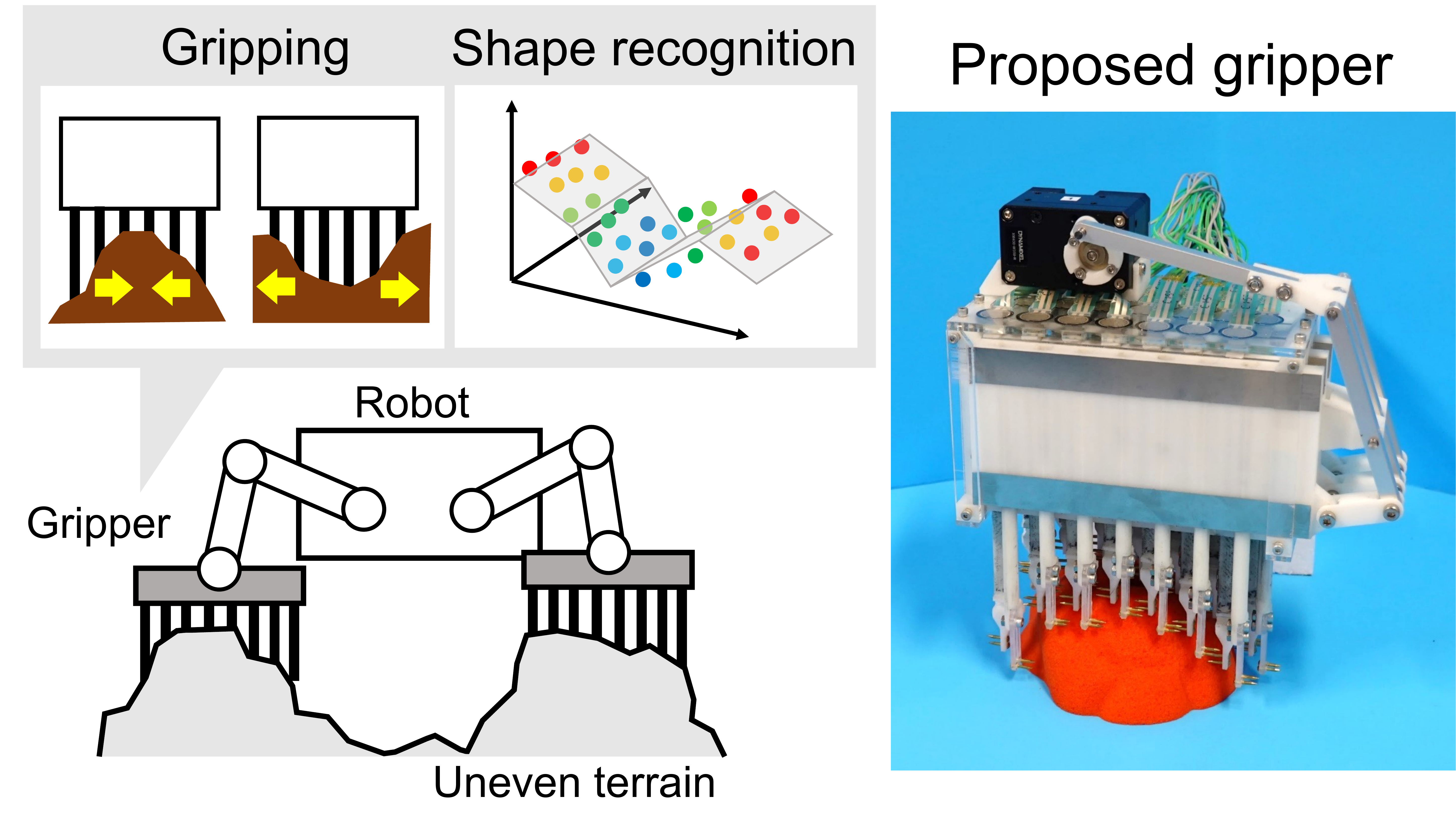}
  \vspace{0mm}\caption{The proposed gripper and conceptual illustration of application to a robot.}\label{fig:proposal_illustration}
\vspace{-2mm}
\end{figure}
\indent
However, these robotic climbing has only been demonstrated in simulated environments composed of easily graspable footholds that are discretely arranged. In contrast, natural terrains extend continuously, and some parts are convex and some are concave. Therefore, robots should observe the terrain and accurately detect the graspable shapes in it. Through this sensing and detection process, the graspable points are randomly extracted on the terrain surface. Graspable target detection using the 3D point cloud of the terrain is also reported \cite{uno_isairas}, however, vision-based terrain sensing and mapping capabilities depend on optical conditions and are limited in terms of accuracy. Moreover, even though the robot can sense the terrain precisely and detect the graspable shapes, the more limited shapes the gripper can grasp, the less graspable areas are on the terrain. Then if the nearest graspable points are not in the range that the robot's feet reach, it will get stuck. Various grippers to grasp the natural terrains have been proposed so far, however, the graspable shapes are limited (e.g., convex, planar only)\cite{dense_array,tomarigi}. As another approach to maximize the graspable parts in terrain, NASA/JPL's quadrupedal rock-climbing robot: LEMUR3's microspine grippers have hundreds of hooks. This redundant design enables it to grasp almost everywhere on the rocky surface. However, this concept results in the large size and weight of the grippers, which made the locomotion slow (three minutes per grip/ungrip cycle). In summary, significant improvements are essential in 1) terrain sensing/mapping and 2) versatile grasping for the climbing robots to perform autonomous locomotion.\\
\indent
In this paper, to address the aforementioned technical problems: limitation of 1) the accuracy in the terrain mapping and 2) the graspability of the various shapes, we propose a novel pin-array gripper that can grasp both convex and concave terrain surfaces and simultaneously measure the shape of the terrain by the tactile sensing (Fig.~\ref{fig:proposal_illustration}). By gripping the terrain surface and measuring its shape at the same time, the robot can move on the terrain surface without any pre-observation processes. We first present the pin-array gripper design that enables simultaneous grasping and shape recognition of the unstructured terrain, as well as the mechanistic model of gripping. We then experimentally evaluate the gripping performance of the prototyped gripper and its functionality. Lastly, we demonstrate 3D terrain mapping with the proposed gripper and discuss the applicability of the system to a mobile robot.
\section{GRIPPER DESIGN}
\subsection{Mechanism for Gripping}
Some flexible grippers have been proposed that adapt and deform themselves to the shape of a target object. This adaptability allows them to grasp universal shapes without posture estimation of the object or grasp planning. However, there are not many grippers capable of a strong grip both convex and concave terrains. Although one example of the universal gripper uses the jamming transition phenomenon of powders or granules\cite{jamming_gripper}, it is hard to apply in a natural environment where the gripper's material property changes. Pin-array mechanisms, such as a contour gauge, have also been proposed. The principal example is the Omnigripper\cite{omnigripper}, which can only grasp convex shapes. Grippers that grasp various objects have been developed by improving the pins' movement and geometry\cite{ctsa_hand,tsinghua_univ}. However, they do not have sufficient power to support a robot because they grasp the target object by friction between pins and the target. Noh \it{et al.} \rm proposed a robot with pin-array legs\cite{pinbot}, but it cannot maintain the robot body stably on inclined surfaces because it does not have grasping ability.

\begin{figure}[t]
\renewcommand{\baselinestretch}{0.8}
\vspace{-1mm}
  \centering
  \includegraphics[width=0.95\linewidth]{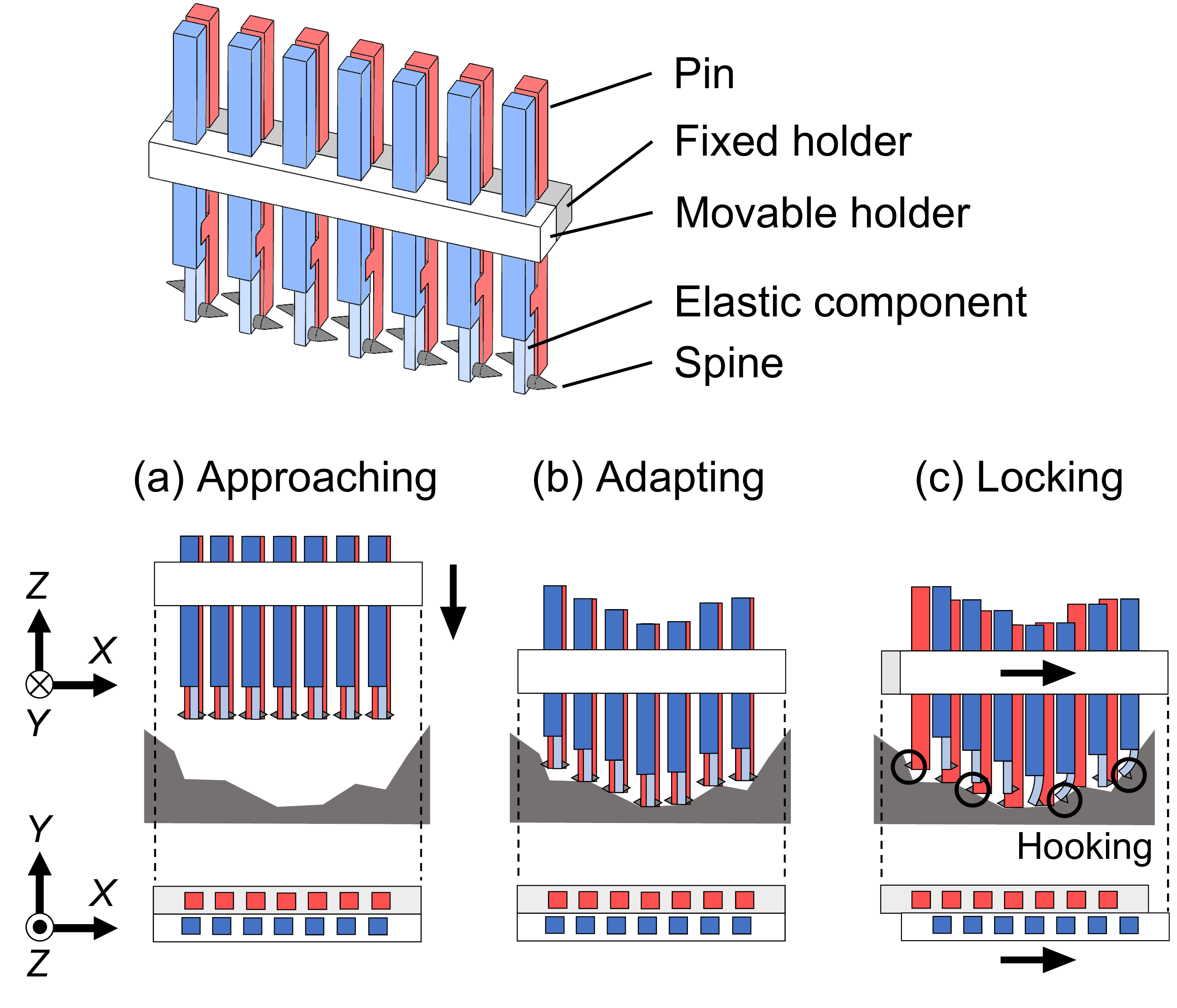}
  \vspace{0mm}\caption{Schematic of the proposed gripping structure. The upper illustration shows the basic structure of the gripper. The lower side view illustrations show the movement of pins and holders when gripping convex/concave shapes.}\label{fig:schematic_diagram}
\vspace{-2mm}
\end{figure}

We propose a new gripping mechanism using a pin-array with spines and two types of holders. The schematic diagram is shown on the upside of Fig.~\ref{fig:schematic_diagram}. The pins are divided vertically into two parts and stored in separate holders. The backside holder is fixed to the gripper, and the front holder can slide horizontally. Each tip of the pin has an elastic component and spines. The elastic force presses the spine against a target to obtain a holding force.\\
\indent
As shown at the bottom in Fig.~\ref{fig:schematic_diagram}, the grasping flow is classified into three phases. In (a) Approaching phase, detailed shape and posture information of a target terrain surface is not required. In (b) Adapting phase, the gripper is simply pressed against a target terrain surface, and the pin-array is passively adapted. Finally, in (c) Locking phase, the movable holder is actively actuated horizontally. This motion enables the spines to press against a target terrain surface, providing a holding force by hooking. The same sequence is used to grasp an object, whether it is convex or concave. For example, when gripping a convex shape, the pins apply a pinching force from the outside, and when gripping a concave shape, they apply an expanding force from the inside. In (c) Locking phase, the vertical movement of all pins should be locked.


\subsection{Mechanistic Model with External Force}
\begin{figure}[t]
\renewcommand{\baselinestretch}{0.8}
\vspace{-1mm}
  \centering
  \includegraphics[width=0.95\linewidth]{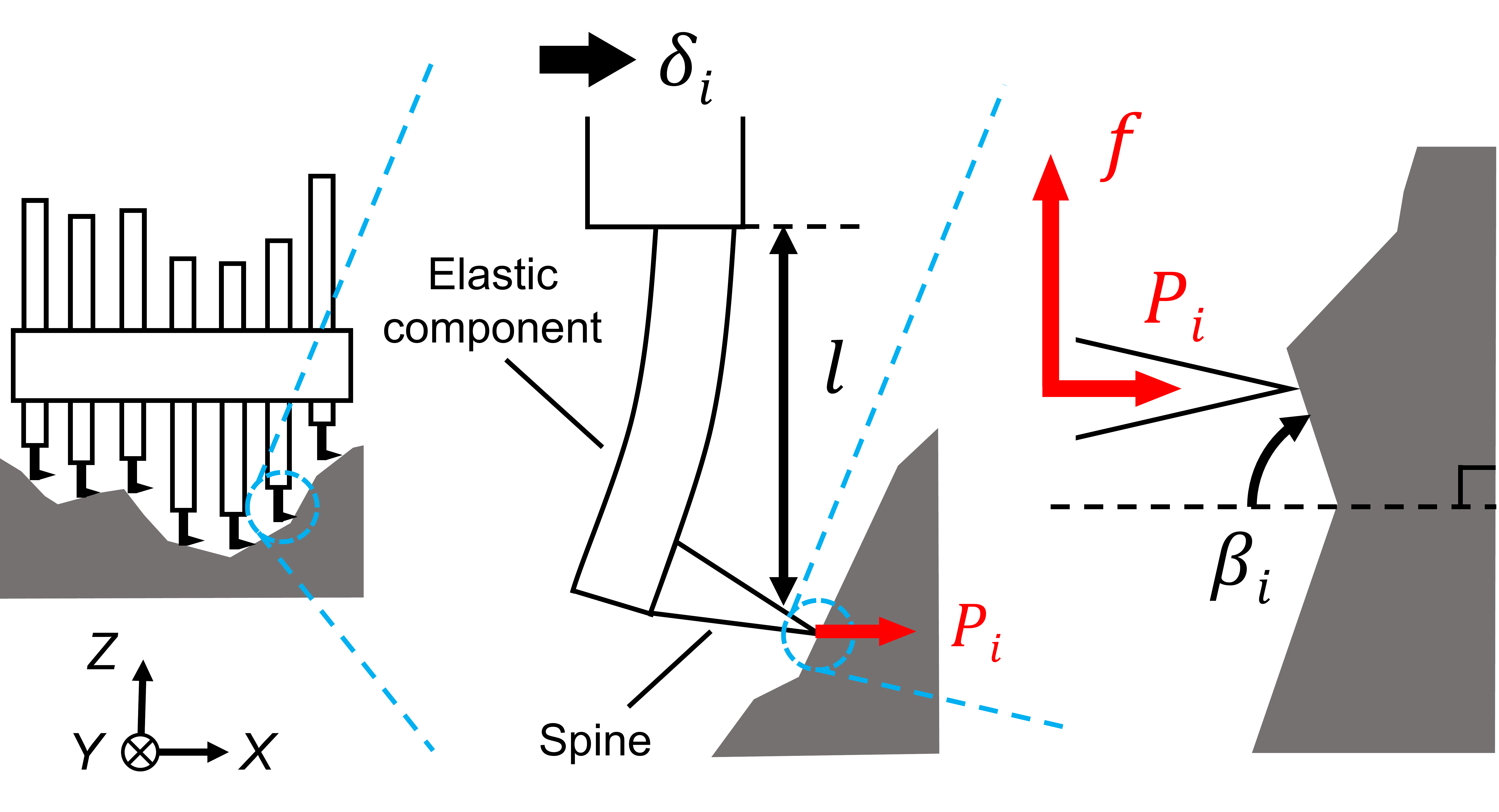}
  \vspace{0mm}\caption{Mechanistic model of the proposed gripper when gripping terrain surface.}\label{fig:mechanistic_model}
\vspace{-2mm}
\end{figure}

In this section, we study the holding force using a mechanistic model when a pulling force is applied to the gripper in the $Z$-axis (vertical) direction. 
The elastic component of the pin can be assumed to be a cantilevered beam, as shown in Fig.~\ref{fig:mechanistic_model}. If the distance that the holder slides after a spine touch a terrain surface is $\delta_i$, the pushing force by the spine $P_i$ is given by
\begin{eqnarray}
P_i = \frac{3\delta_iEI}{l^3}\label{eq:1}
\end{eqnarray}
where $E$ is Young's modulus of the elastic component, $I$ is the second moment of area, and $l$ is the distance from the fixed end to the spine. 

Hereafter, we discuss a frictional model between a spine and a terrain surface.
Asbeck \it{et al.} \rm have studied the friction model between a sharp-tip object and rough terrain surface with microscopic asperities\cite{microspine}. Based on this, the asperity can be assumed as a triangular projection, as illustrated on the right side of Fig.~\ref{fig:mechanistic_model}. Given an external force $f$ in the $Z$-axis to per pin, the condition for no-slip is expressed as follows.
\begin{eqnarray}
f < \mu'_iP_i\;\;\;(\mu'_i = \frac{1 + \mu\tan\beta_i}{\tan\beta_i - \mu})\label{eq:3} 
\end{eqnarray}
Where $\mu'_i$ is the local friction coefficient, and $\mu$ is the global friction coefficient between the spine and the terrain surface. $\mu'_i$ depends on the angle $\beta_i$ of the asperity and diverges to infinity as $\beta_i$ approaches 0. In nature, however, if a large external force is applied, the asperity is broken by the spine and slipped. \\
\indent
Assuming that the gripper is applied an external force in the $Z$-axis and it is equally applied to each pin in contact with the terrain surface, the condition that the gripper cannot come off is given by
\begin{eqnarray}
F < 2\sum_{i=1}^n\mu'_iP_i\label{eq:4}
\end{eqnarray}
where $n$ is the number of pins in contact with the terrain surface. In practice, the holding force varies because external forces applied to each pin are not always equal, and they can also break the spines and terrain surface.

\subsection{Recognition of Terrain Shapes}

\begin{figure}[t]
\renewcommand{\baselinestretch}{0.8}
\vspace{-1mm}
  \centering
  \includegraphics[width=0.90\linewidth]{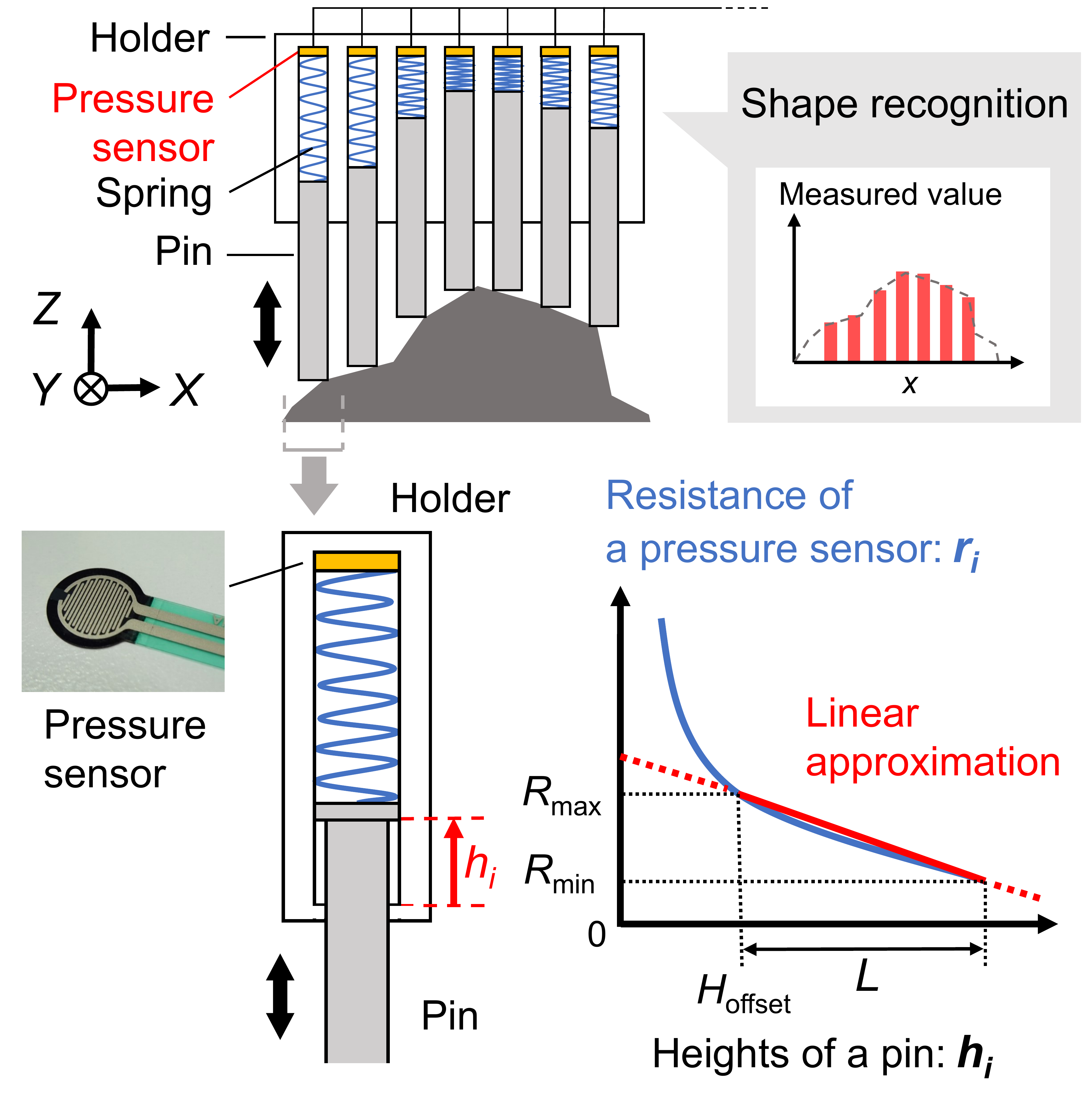}
  \vspace{0mm}\caption{Procedure of shape recognition by the gripper. The lower illustration shows a close-up of the sensor section and the relationship between the height of a pin $h_i$ and the sensor resistance value $r_i$.}\label{fig:shape_recognition}
\vspace{-2mm}
\end{figure}
We introduce the second function of the gripper, that is shape recognition capability by tactile. In natural environments, climbing robots should observe the terrain for path planning and graspable point detection. Although vision sensors such as depth cameras are often applied for the robot's vision, the sensing accuracy depends on the given optical conditions. For example, a passive stereo camera cannot sense in the dark. Active stereo and Time-of-Flight (ToF) cameras are also less accurate when the terrain or the camera is exposed to direct sunlight \cite{uno_jsass}. In addition, while depth cameras and LiDARs have a minimum length for the measurable depth range, it is difficult for a climbing robot to maintain a proper distance between the sensor and the terrain surface because the climbing robot is required to minimize the base height to keep the stability against the tipping over, so the distance to the terrain surface tends to be too close for the sensor. \\
\indent
Tactile sensing can be a solution to address these issues. Several methods have been proposed to classify objects by built-in sensors in a typical robotic gripper. 
Tanaka \it{et al.} \rm proposed a method to classify objects by active touch using a three-finger gripper with tactile sensors on each finger \cite{tanaka_iros2014}. However, when we use a rigid gripper, we can only obtain information at a few points where the object is contacted by the fingers. A method to acquire more information by increasing contact with a soft gripper has been proposed \cite{fingripper}, but the mounted tactile sensors are only two, and high-density shape information has not been obtained.


In contrast, we designed a gripper with a sensor on every pin of the pin-array structure and almost ten times as many sensors as the aforementioned methods to acquire dense shape information that enables terrain recognition. A schematic of the proposed system is shown in Fig.~\ref{fig:shape_recognition}. In the developed gripper prototype, we employed a pressure sensor to measure the pin displacement because of the miniatualized size. By the developed gripper, when a pin moves vertically, the resistance value of the built-in pressure sensor varies by the attached spring. In the proposed system, we obtain the height of the terrain by a linear approximation between the resistance and the height of a pin. When the gripper is pressed against a terrain surface, if the sensor resistance of a pin is $r_i$, the measured height, $h_i$, is given by
\begin{eqnarray}
h_i = \frac{r_i-R_{\rm{max}}}{R_{\rm{min}}-R_{\rm{max}}}+L+H_{\rm{offset}}\label{eq:5}
\end{eqnarray}
where $R_{\rm{max}}$ and $R_{\rm{min}}$ are the maximum and minimum resistance in the range of the linear approximation. $L$ is the distance a pin moves from $R_{\rm{min}}$ to $R_{\rm{max}}$ of the sensor's resistance: displacement measurement range. $H_{\rm{offset}}$ is the distance from a pin at its most extended state to the measurable range.


\section{GRIPPER PROTOTYPE}
\begin{figure}[t]
\renewcommand{\baselinestretch}{0.8}
\vspace{-5mm}
  \centering
  \includegraphics[width=0.80\linewidth]{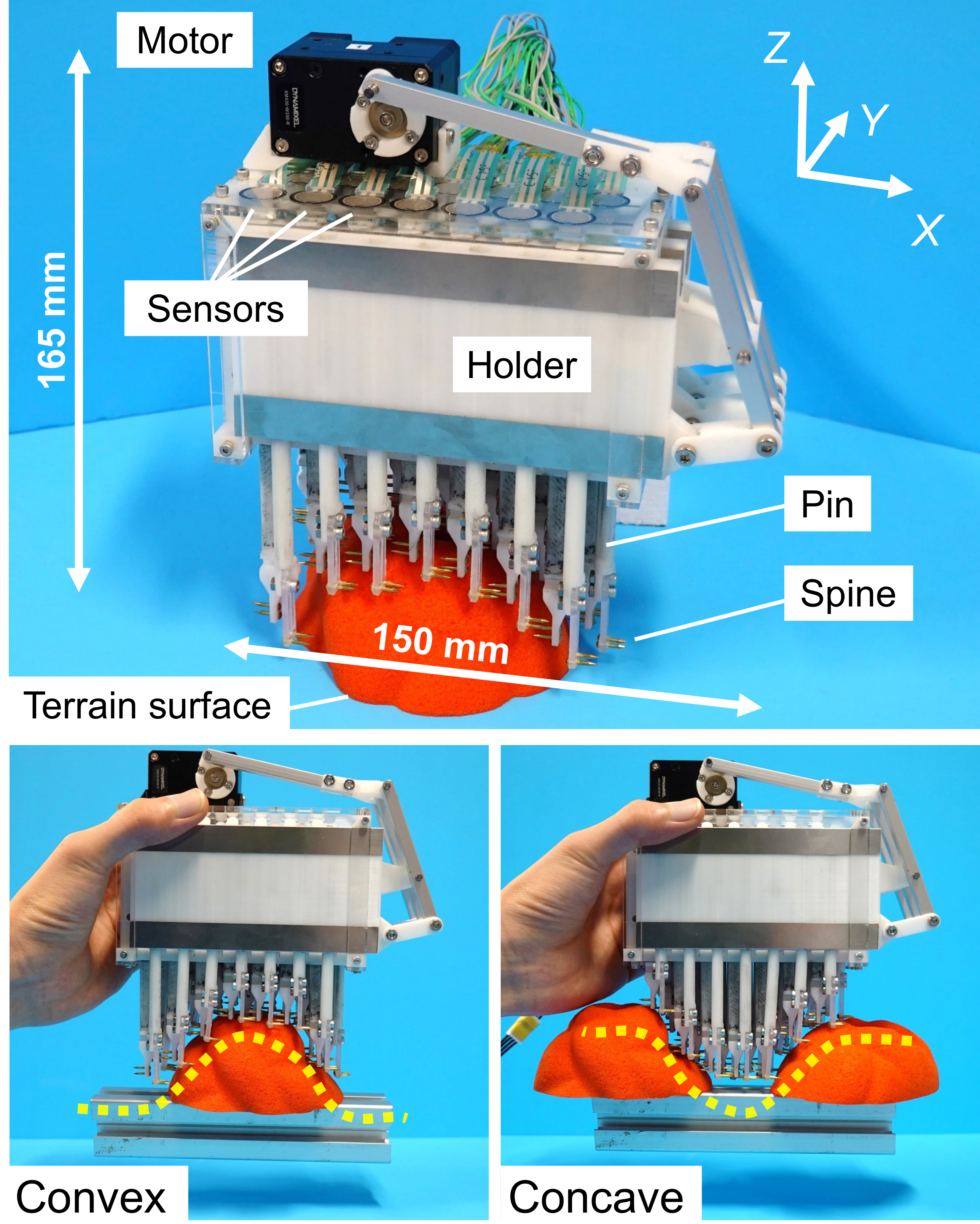}
  \vspace{0mm}\caption{Overview of the gripper prototype (top), gripping and lifting convex/concave terrains composed of climbing holds (bottom).}\label{fig:gripper_prototype}
\vspace{-1mm}
\end{figure}

We developed a prototype of the gripper shown at the top of Fig.~\ref{fig:gripper_prototype}. The width of the prototype is 150~mm, the height is 165~mm (when the pins are extended), the depth is 70~mm, and the whole mass is 522~g. It consists of three blocks, and each of them is composed of seven pins and two holders. 
The distances between pins are $X_{\rm{pitch}}=14~\rm{mm}$ along the $X$-axis and $Y_{\rm{pitch}}=17.4~\rm{mm}$ along the $Y$-axis. The holders and pins are made of ABS plastic, the elastic components are made of polycarbonate, and the spines are brass nails.

\begin{figure}[t]
\renewcommand{\baselinestretch}{0.8}
\vspace{-1mm}
  \centering
  \includegraphics[width=0.95\linewidth]{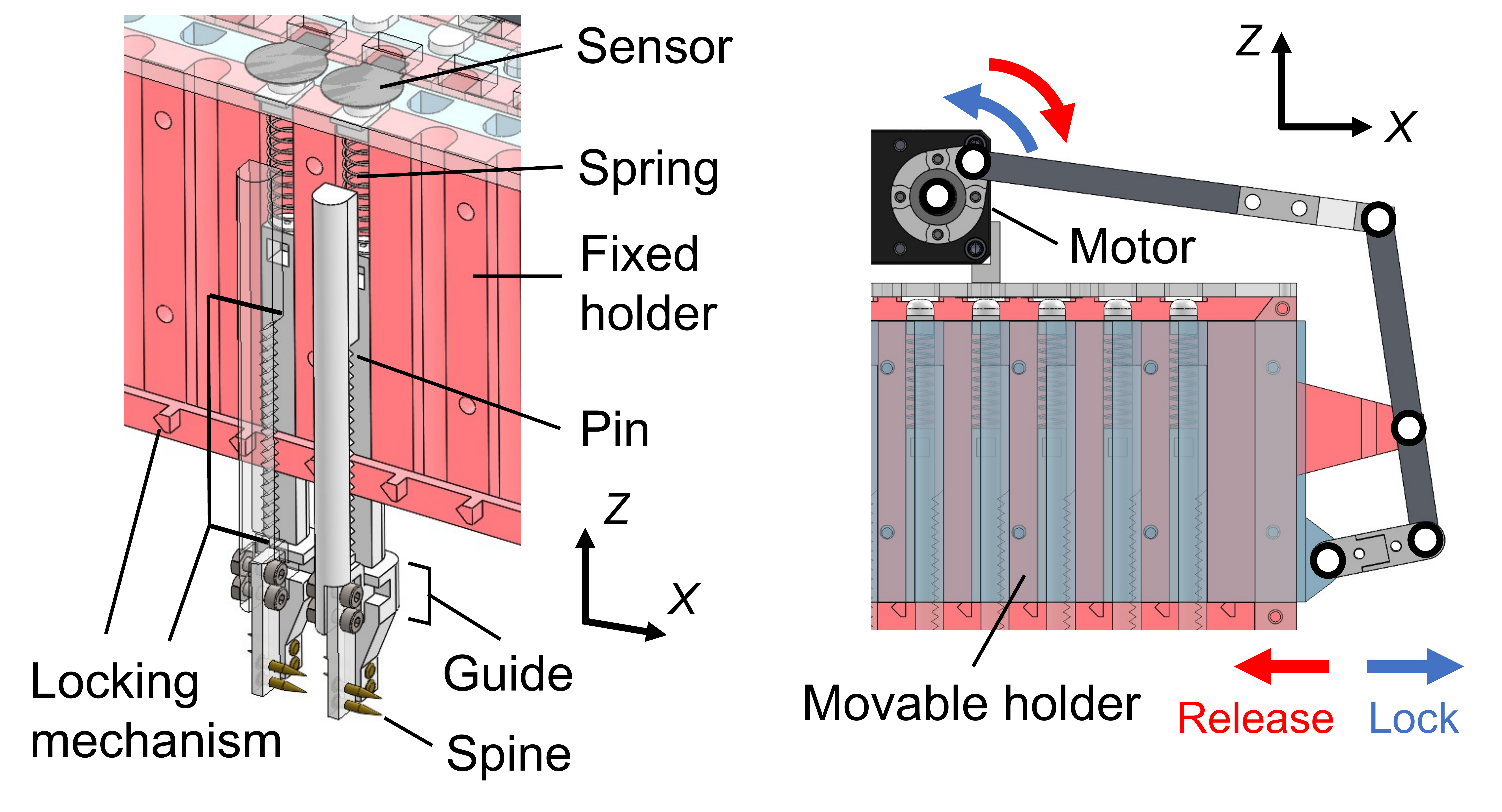}
  \vspace{0mm}\caption{Close-up of the pin section (left, several pins and a movable holder are omitted for visibility). Link mechanism to actuate movable holders (right).}\label{fig:pin}
\vspace{-2mm}
\end{figure}

The close-up of the pin section is shown on the left side of Fig.~\ref{fig:pin}. At the bottom of the pins, there are guides to prevent the vertically divided pins from moving each other along the $Z$-axis. On the top, there are pressure sensors and springs for shape recognition. The locking mechanisms on fixed holders and pins restrain the $Z$-axis movement of all pins at the (c) Locking phase. The holders are actuated by a servo motor and a link mechanism.
The right side of Fig.~\ref{fig:pin} shows the motion of the link mechanism that translates the motor rotation into the linear motion of the holders.
It allows the gripper to grasp/release by simply turning the motor at the same angle, regardless of the terrain shapes.\\
\indent
We tested the prototype on an emulated terrain made of climbing holds as shown at the bottom of Fig.~\ref{fig:gripper_prototype}. We confirmed that it could hold a 500~g object with both convex and concave shapes. In addition, we evaluated the pressure sensors. We examined the relationship between pin heights and sensor resistances for every 21~pins. From the results, we determined the linearized range as in Fig.~\ref{fig:shape_recognition}, where $L=20~\rm{mm}$ and $H_{\rm{offset}}=16~\rm{mm}$. $R_{\rm{max}}$ and $R_{\rm{min}}$ are determined individually for each pin since they vary depending on the individual sensor.



\begin{figure}[t]
\renewcommand{\baselinestretch}{0.8}
\vspace{-1mm}
  \centering
  \includegraphics[width=0.80\linewidth]{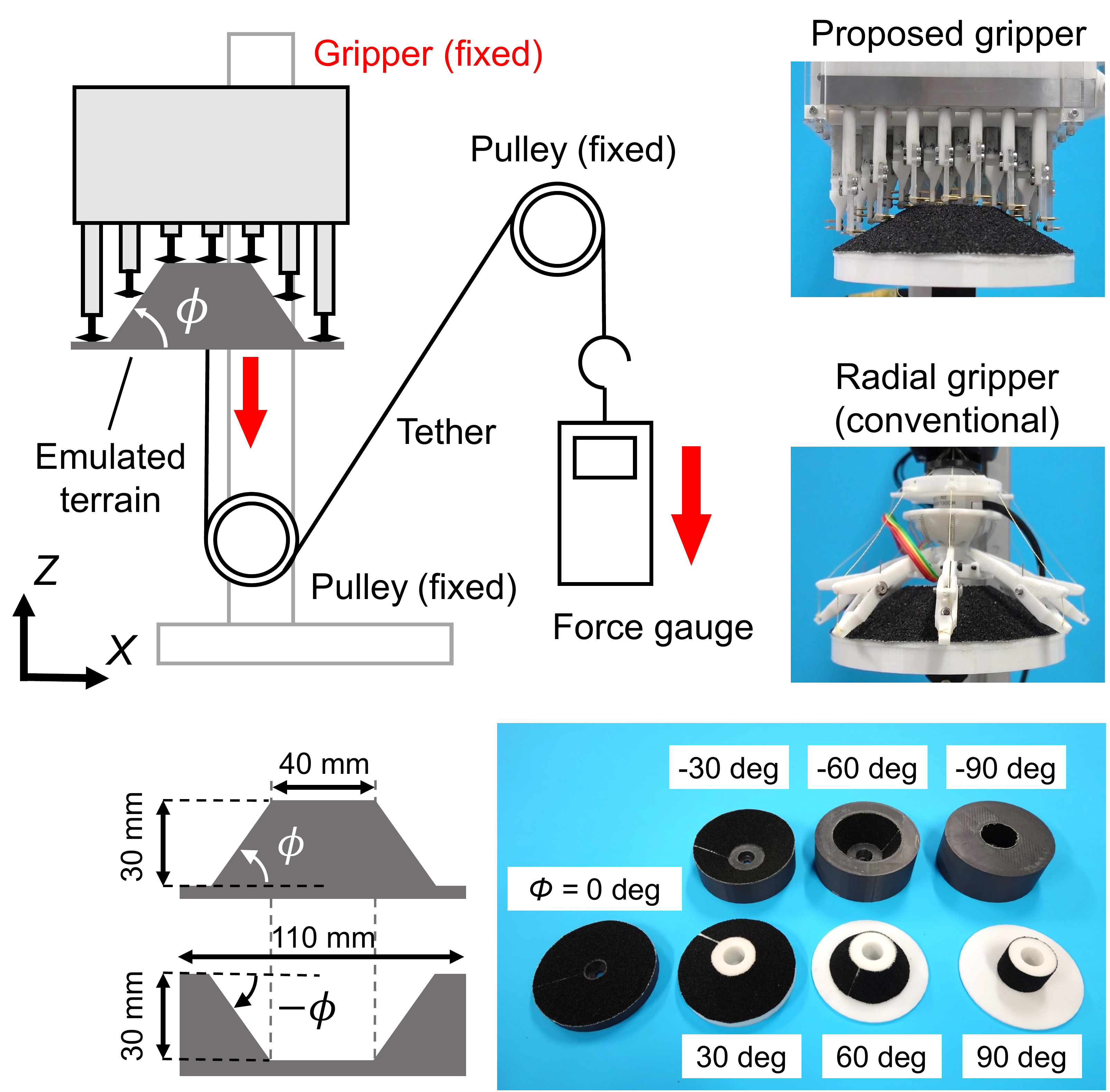}
  \vspace{0mm}\caption{Experimental setup to evaluate the gripping performance in different shapes of the gripped target (top), emulated convex and concave terrain shapes used in the experiment (bottom).}\label{fig:gripping_experiment}
\vspace{-2mm}
\end{figure}

\section{EVALUATION}
\subsection{Gripping Performance for Various Shapes}
We first evaluated the gripping performance of convex and concave terrain shapes using the prototype. In addition, the same experiment was conducted using a conventional gripper and the results were compared.\\
\indent
An overview of the experiments is shown at the top of Fig.~\ref{fig:gripping_experiment}, and the emulated terrains are shown at the bottom. The gripper fixed to the experimental device gripped an emulated terrain and applied an external pulling force in the negative direction of the $Z$-axis. The pulling force was measured with a force gauge until the emulated terrain was released. The holding force was defined as the maximum pulling force plus the gravity applied to the emulated terrains. If the gripper failed to grasp and the emulated terrain fell, the holding force was defined as 0~N. The emulated terrains were seven different types with inclination angles $\phi$, each covered with \#40 sandpaper. Measurements were taken 10 times on each emulated terrain. The conventional gripper used for comparison has eight fingers with spines attached in a radial pattern, the same finger arrangement used in HubRobo\cite{uno_humanoids}.


\begin{figure}[t]
\renewcommand{\baselinestretch}{0.8}
\vspace{-1mm}
  \centering
  \includegraphics[width=.8\linewidth]{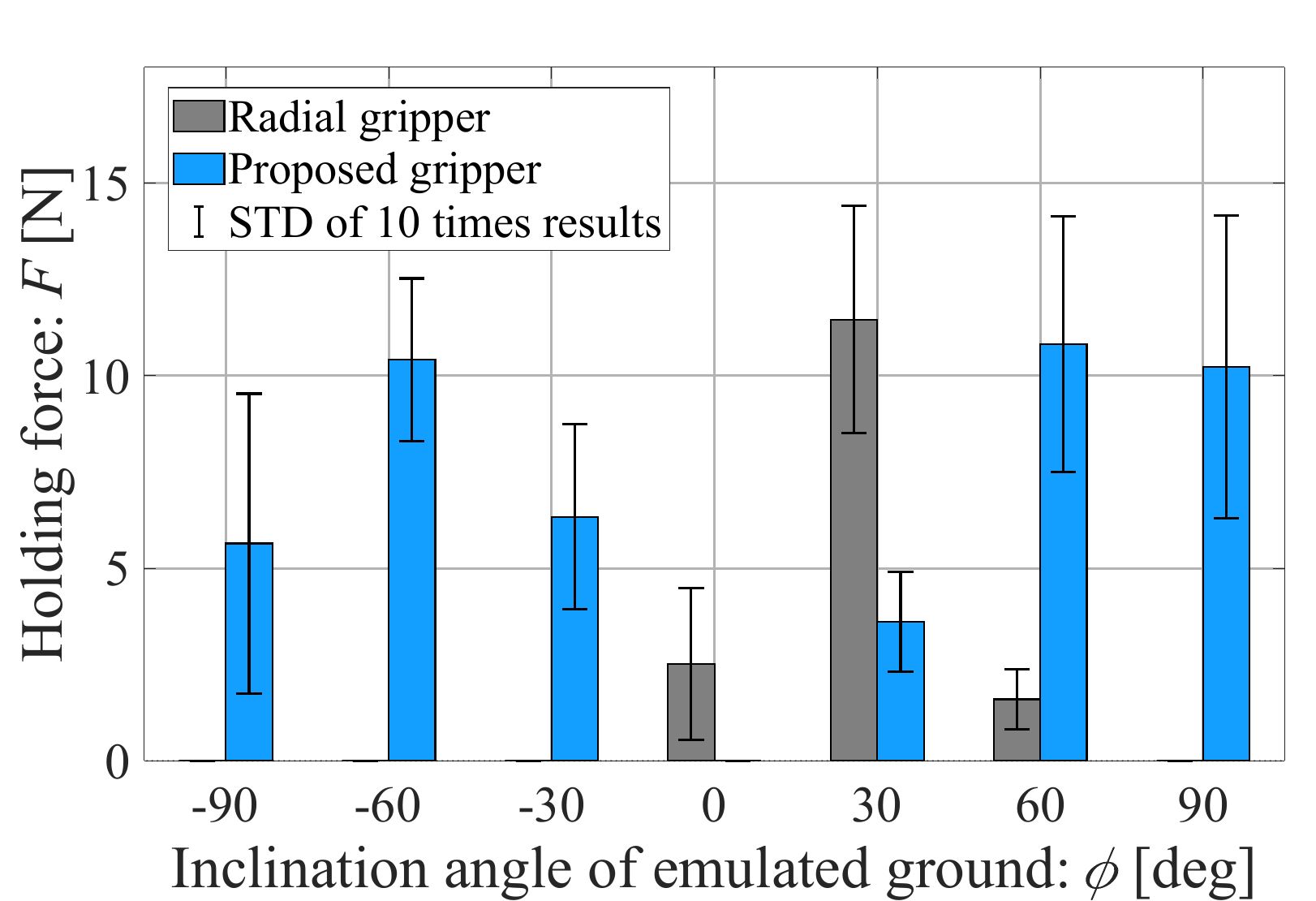}
  \vspace{0mm}\caption{Holding force measured by the force gauge for each inclination angle of emulated terrain shapes. Error bars are the standard deviation (STD) of holding force from ten times results for each $\phi$.}\label{fig:gripping_result}
\vspace{-2mm}
\end{figure}

As a result of the experiment, the relationship between the emulated terrain's inclination angle $\phi$ and the average value of holding force $F$ is shown in Fig.~\ref{fig:gripping_result}. Error bars are the standard deviation (STD) of holding forces. When the inclination angle $\phi$ is negative, the terrain shape is concave, and when it is positive, the shape is convex. The proposed gripper obtained a steady holding force of 5\textasciitilde10~N for more shapes than conventional grippers. 
The holding force decreased as the inclination became more gentle in both cases convex and concave. It shows that the angle of microscopic asperity $\beta_i$ increases as the slope becomes more gentle, and the local friction coefficient $\mu'_i$ becomes smaller, as shown in the equation (\ref{eq:3}).
\indent
At $\phi=90~\rm{deg}$ and $\phi=-90~\rm{deg}$, the holding force was smaller, and the variability was larger than at $\phi=60~\rm{deg}$ and $\phi=-60~\rm{deg}$. It indicates that the surface area of the emulated terrain was smaller, and the number of pins $n$ whose spines touched the emulated terrain was reduced, as shown in equation (\ref{eq:4}). Since $n$ was determined randomly, the smaller surface area had a greater effect on the variation in holding force.

\subsection{Shape Recognition}
\begin{figure}[t]
\renewcommand{\baselinestretch}{0.8}
\vspace{-1mm}
  \centering
  \includegraphics[width=0.85\linewidth]{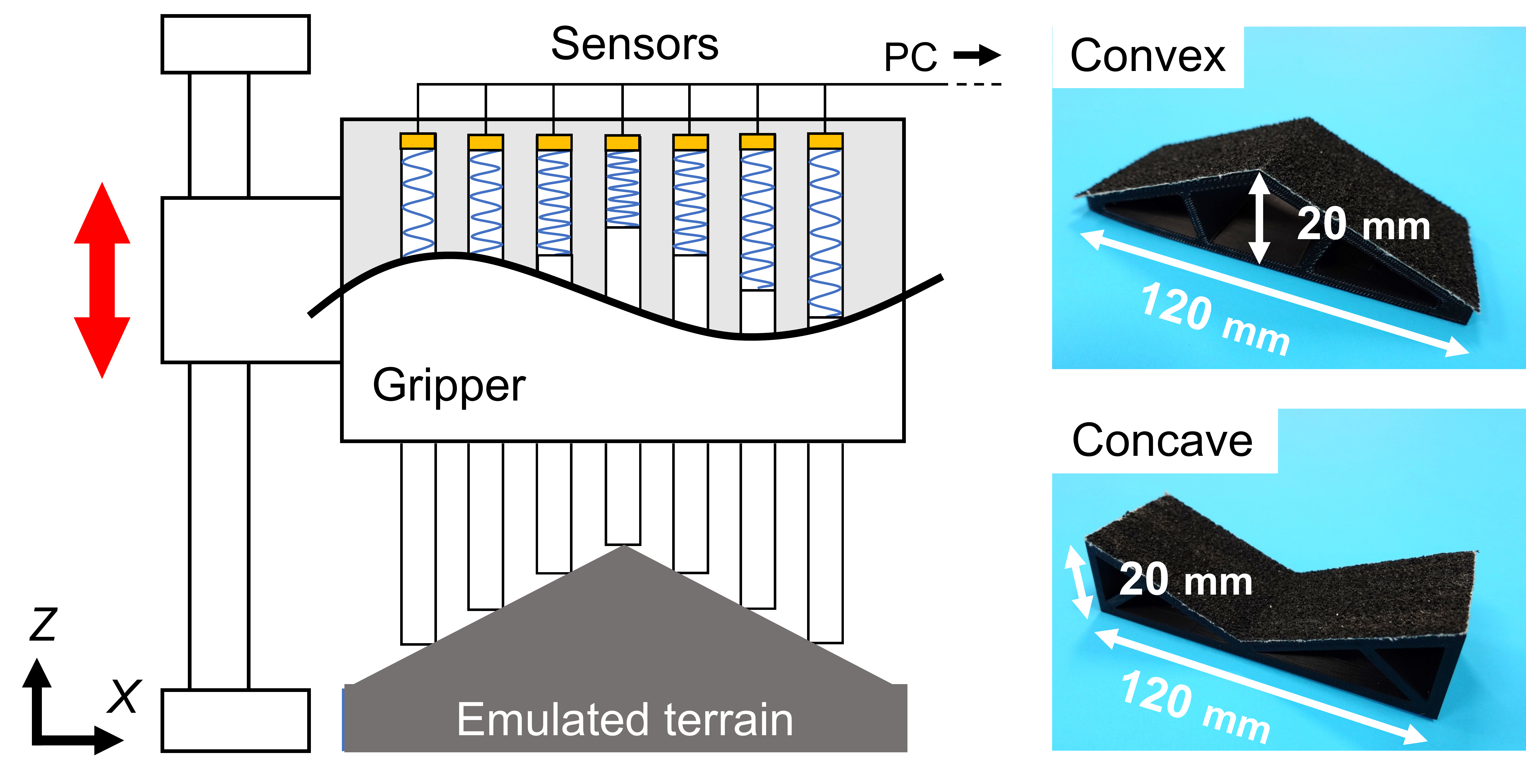}
  \vspace{0mm}\caption{Experimental setup to evaluate shape recognition and emulated grounds.}\label{fig:shape_experiment}
\vspace{-2mm}
\end{figure}

Shape recognition by the proposed gripper was evaluated using the prototype. We examined the accuracy of the measured shapes by comparing them with ground truth.\\
\indent
An overview of the experiment shows in Fig.~\ref{fig:shape_experiment}. The emulated terrains were two types: convex and concave with heights of 20~mm and flat in the depth direction. The gripper was pressed against the terrain surface by the device until the pins were sufficiently adapted. After that, we measured the heights of each pin by sensors in the gripper. The heights of pins are determined from the resistance values of the pressure sensors, as shown in equation (\ref{eq:5}). 
We measured 10 times on each emulated terrain.

\begin{figure}[t]
\renewcommand{\baselinestretch}{0.8}
\vspace{-1mm}
  \centering
  \includegraphics[width=0.85\linewidth]{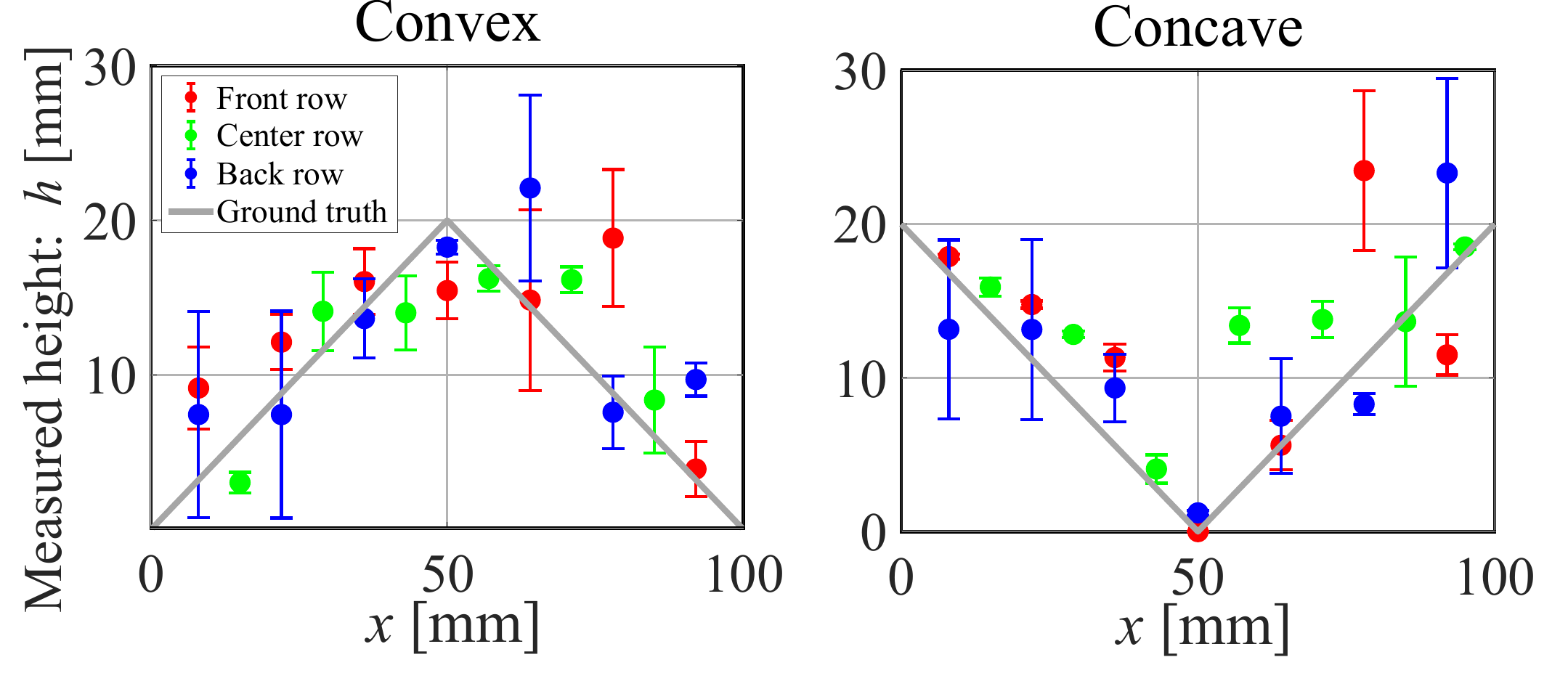}
  \vspace{0mm}\caption{Measured height of the emulated terrain by the gripper for each convex and concave terrain shape. Error bars are the standard deviation (STD) of height in the 10 times experiment.}\label{fig:shape_result}
\vspace{-2mm}
\end{figure}

Fig.~\ref{fig:shape_result} shows the average measured heights for each emulated terrain. The gray lines show the ground truth. Error bars are the standard deviation of measured heights. We could obtain shape information indicating the convexity and concavity characteristics of the terrains. 

Standard deviations of the measured values at each emulated terrain are $\bar{\sigma}_{\rm{convex}}=2.71~\rm{mm}$ and $\bar{\sigma}_{\rm{concave}}=2.02~\rm{mm}$ for the average of all pins. However, the standard deviation varies significantly among individual sensors, as shown in the graph. The maximum value of the standard deviation is $\sigma_{\rm{max}}=6.69~\rm{mm}$, and the minimum is $\sigma_{\rm{min}}=0.16~\rm{mm}$. In order to achieve accurate shape recognition, it is required to reduce the variation. 
\indent
These results show that the proposed system provides high-density information on shape and size by simply pressing the gripper against the terrain surface. The system is expected to generate a physical interaction with the unknown environment by simultaneously gripping and recognizing the terrain, thereby improving the robot's autonomous locomotion.

\subsection{3D Mapping}
\begin{figure}[t]
\renewcommand{\baselinestretch}{0.8}
\vspace{-1mm}
  \centering
  \includegraphics[width=0.85\linewidth]{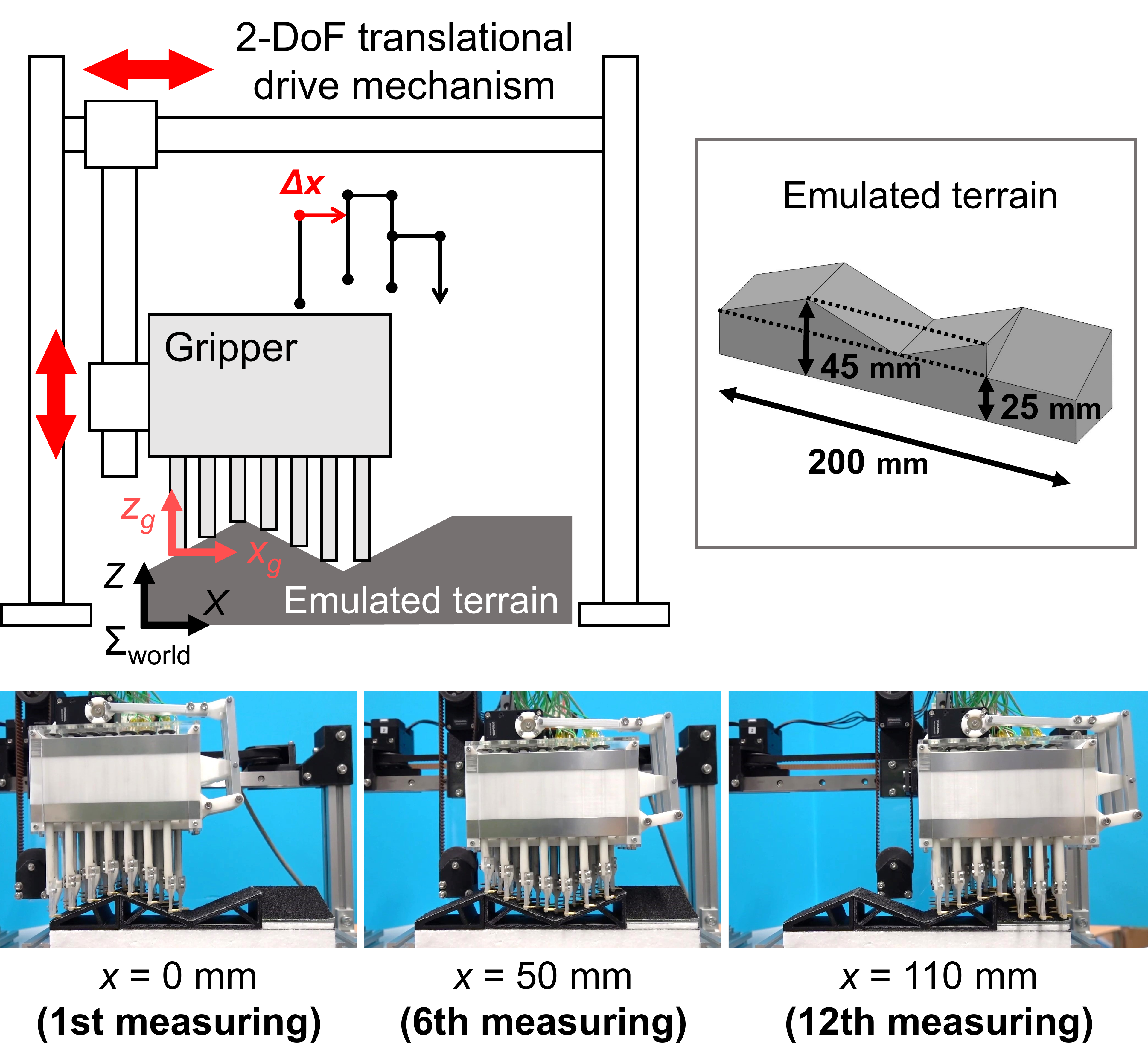}
  \vspace{0mm}\caption{Experimental setup to evaluate 3D mapping by the gripper and the emulated terrain.}\label{fig:mapping_experiment}
\vspace{0mm}
\end{figure}

It is necessary for a mobile robot in an unknown environment to map its surroundings for action planning of the mobile robot. We experimented with 3D mapping of uneven terrain by combining shape recognition and gripper location information. The accuracy of the mapping was evaluated by comparison with ground truth.\\
\indent
The experimental setup consists of the gripper and the machine that actuates the gripper, as shown in Fig.~\ref{fig:mapping_experiment}. The machine is a two degree-of-freedom translational drive mechanism that can move the gripper to any position in the $x$--$z$ plane. The mapping is performed using the machine and the gripper by repeating the following actions.

\begin{figure}[t]
\renewcommand{\baselinestretch}{0.8}
\vspace{-1mm}
  \centering
  \includegraphics[width=0.80\linewidth]{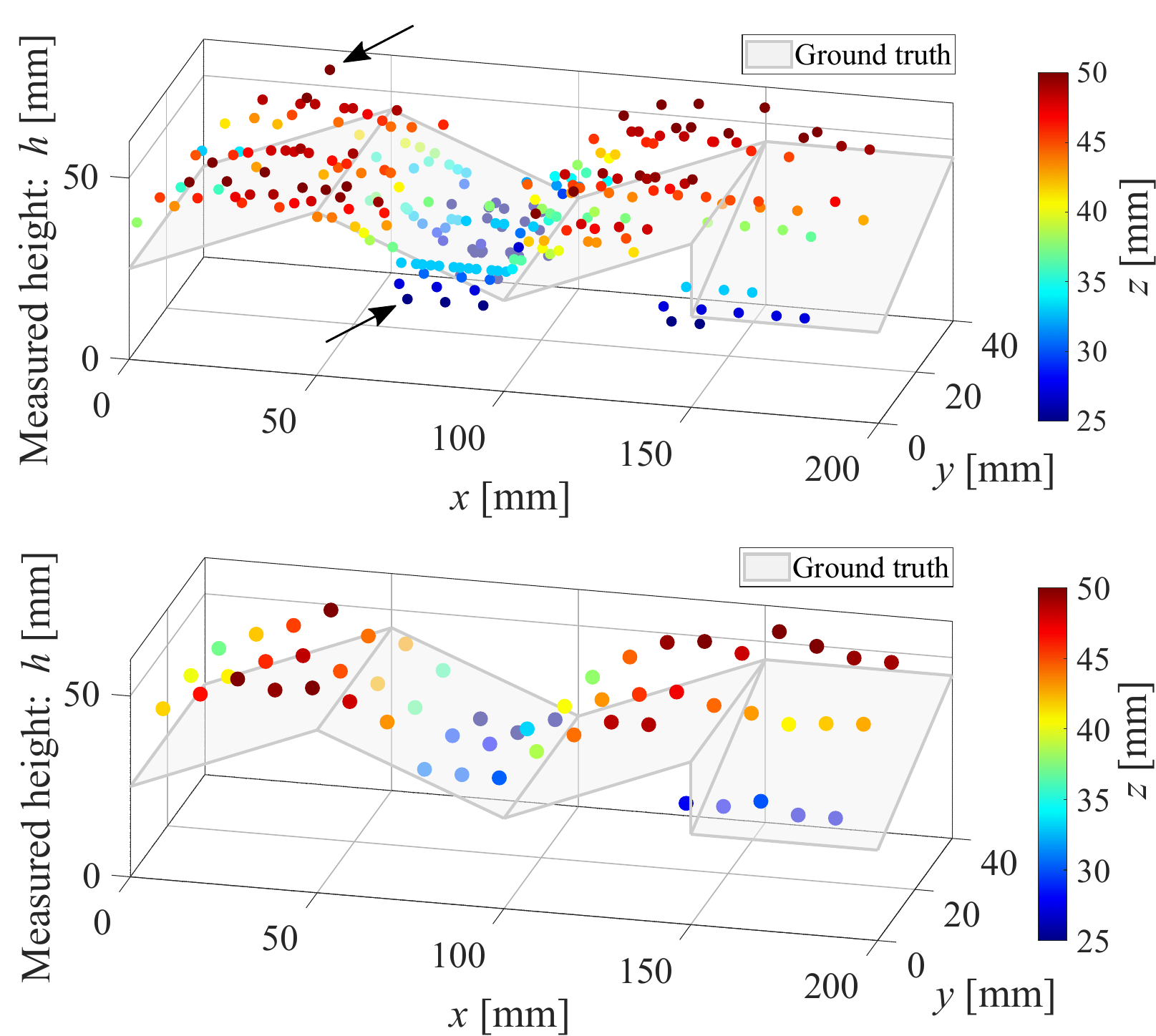}
  \vspace{0mm}\caption{3D map generated by the gripper. The upper figure shows all the point clouds acquired by the gripper. The lower shows averages of the $z$ values of the separated point cloud by square columns on the $x$--$y$ plane. The points far off the ground truth in the upside graph (marked by arrows) were removed by the averaging.}\label{fig:mapping_result}
\vspace{-2mm}
\end{figure}

\begin{enumerate}
\item Move the gripper by $\Delta x$ along the $X$-axis.
\item Move the gripper in the negative direction of the $Z$-axis and press against the emulated terrain surface until the pins are sufficiently adapted.
\item Get the value of sensors in the gripper and the gripper's current position.
\item Move the gripper in the positive direction of the $Z$-axis until all pins leave the terrain surface.
\end{enumerate}
We set $\Delta x=10~\rm{mm}$ and repeated the above operation 12 times, then generated a point cloud from all the sensor values and current gripper positions. 
Given the gripper position $(x_g, z_g)$, each coordinate of the point cloud in the World coordinate system is expressed as follows.
\begin{eqnarray}
x_i=x_g+jX_{\rm{pitch}},\;\;y_i=kY_{\rm{pitch}},\;\;z_i=z_g+h_i \label{eq:6}
\end{eqnarray}
where $j$ and $k$ are numbers representing the position of a pin ($j=1, 2, ...,7$, $k=1, 2, 3$). $X_{\rm{pitch}}$ and $Y_{\rm{pitch}}$ are distance between pins in each direction. $h_i$ is the measured height of each pin by built-in sensors, as shown in equation (\ref{eq:5}).
The width of the emulated terrain is 200~mm, the height is 45~mm, and the depth is 40~mm as shown in Fig.~\ref{fig:mapping_experiment}. 
It includes convex, concave, and a slope at depth, with a height displacement of 20~mm.\\
\indent
The point cloud obtained by the experiment is shown in Fig.~\ref{fig:mapping_result}. The upper figure shows all the point clouds obtained from the measurements. The lower shows averages of the $z$ values of the separated point cloud by square columns of $x=10~\rm{mm}$ and $y=15~\rm{mm}$. The gray plane is the ground truth. The average height error from the ground truth at each point in the lower figure was $\bar{e}=\rm{7.67~mm}$.\\
\indent
The proposed system could reconstruct the map of the convex and concave shapes in the emulated terrain as a point cloud. The slope along the $Y$-axis was successfully mapped as well. Furthermore, the points far off the ground truth in the upside graph (marked by arrows) were removed by averaging with the neighboring points. It indicates that by measuring the same location repeatedly with different pins, the individual differences between sensors were an issue in Section IV-B, are offset, and accurate information is obtained.

\section{CONCLUSION}
In this paper, we presented a pin-array gripper capable of gripping various shaped terrain surfaces and measuring the shape at the same time. We prototyped the gripper and experimentally evaluated its grasping and sensing functions. The proposed gripper is featured in the following points: 1) versatile graspability: the proposed gripper can grip more various shapes, including concaves, than the conventional gripper without any preobservation or grasping planning; 2) terrain measuring capability: we experimentally demonstrated the feasibility of terrain shape recognition (convex or concave) by the proposed gripper. In addition, we applied the function to a 3D terrain mapping application and confirmed its effectiveness. In conclusion, the proposed gripper has a sufficient capability of versatile grasping and accurate terrain sensing/mapping simultaneously, thus the proposed system is effective for mobile robots in unstructured environments to move autonomously and reliably.\\ 
\indent
In future work, the authors plan to develop a robot with the proposed gripper installed on its limbs and conduct field test campaigns on a cliff to confirm its effectiveness in more practical situations. 
Improvements to the sensing system of the gripper are also following this work to obtain more dense shape and force information by combining different sensors such as force/torque sensors and vision sensors.

\addtolength{\textheight}{-12cm}   




\begin{thebibliography}{99}

\bibitem{brooks}
R. A. Brooks,
{\rm ``New Approaches to Robotics'',} \textit{Science}\rm, vol.~253, issue~5025, pp.~1227--1232, 1991.

\bibitem{pfeifer}
R. Pfeifer, \textit{et al.},
{\rm ``Morphological Computation – Connecting Brain, Body, and Environment'',} \textit{Jpn. Sci. Monthly}\rm, vol.~58, no.~2, pp.~48--54, 2005.

\bibitem{wild_anymal}
T. {Miki}, \textit{et al.},
{\rm ``Learning Robust Perceptive Locomotion for Quadrupedal Robots in the Wild''}\rm, \textit{Sci. Robot.}\rm, vol.~7, no.~62, 2022.

\bibitem{yoshida2002} 
K. {Yoshida}, \textit{et al.},
{\rm ``A Novel Strategy for Asteroid Exploration with a Surface Robot'', }\rm \textit{Proc. 34th COSPAR Scientific Assembly}\rm, 2002.

\bibitem{scaler}
Y. Tanaka, \textit{et al.},
{\rm ``SCALER: A Tough Versatile Quadruped Free-Climber Robot'', }\rm \textit{Proc. IEEE/RSJ Int. Conf. Intel. Rob. Syst.}\rm, 2022.

\bibitem{uno_humanoids}
K. Uno, \textit{et al.},
{\rm ``Hubrobo: A Lightweight Multi-Limbed Climbing Robot for Exploration in Challenging Terrain'', }\rm \textit{Proc. IEEE-RAS 20th Int. Conf. Humanoid Rob.}\rm, pp.~209--215, 2021.

\bibitem{shirai_iros}
Y. Shirai, \textit{et al.},
{\rm ``Simultaneous Contact-Rich Grasping and Locomotion via Distributed Optimization Enabling Free-Climbing for Multi-Limbed Robots'', }\rm \textit{Proc. IEEE/RSJ Int. Conf. Intel. Rob. Syst.}\rm, 2022.

\bibitem{nagaoka_ral}
K. Nagaoka, \textit{et al.},
{\rm ``Passive Spine Gripper for Free-Climbing Robot in Extreme Terrain'', }\textit{IEEE Robot. Automat. Lett.}\rm, vol.~3, no.~3, pp.~1765--1770, 2018.


\bibitem{uno_isairas}
K. Uno \textit{et al.},
{\rm ``Non-Periodic Gait Planning Based on Salient Region Detection for a Planetary Cave Exploration Robot'',} \textit{Proc. Int. Symp. Artif. Intell., Robot., Automat. Space}\rm, no.~5027, 2020.


\bibitem{dense_array}
S. Wang, \textit{et al.},
{\rm ``A Palm for a Rock Climbing Robot Based on Dense Arrays of Micro-Spines'',} \textit{Proc. IEEE/RSJ Int. Conf. Intel. Rob. Syst.}\rm, pp.~52--59, 2016.


\bibitem{tomarigi}
W. R. T. Roderick, \textit{et al.},
{\rm ``Bird-Inspired Dynamic
Grasping and Perching in Arboreal Environments'',} \textit{Sci. Robot.}, vol.~6, no.~61, 2021.

\bibitem{lemur3}
A. Parness, \textit{et al.},
{\rm ``LEMUR 3: A Limbed Climbing Robot for Extreme Terrain Mobility in Space'', }\rm \textit{Proc. IEEE Int. Conf. Robot. Automat.}\rm, pp.~5467--5473, 2017.

\bibitem{jamming_gripper}
E. Broun, \textit{et al.},
{\rm ``Universal Robotic Gripper Based on the Jamming of Granular Material'',} \textit{Proc. National Academy of Sciences}\rm, vol.~107, no.~44, pp.~18809--18814, 2010.

\bibitem{omnigripper}
P. B. Scott,
{\rm ``The `Omnigripper': A Form of Robot Universal Gripper'',} \textit{Robotica}\rm, vol.~3, issue~3, 1985.

\bibitem{ctsa_hand}
H. Fu, \textit{et al.},
{\rm ``A Novel Cluster-Tube Self-Adaptive Robot Hand'',} \textit{Robot. biomim.}\rm, vol.~4, no.~1, art.~no.~25, 2017.

\bibitem{tsinghua_univ}
A. Mo, \textit{et al.},
{\rm ``A Novel Universal Gripper Based on Meshed Pin Array'',} \textit{Int. J. Advanced Robotic Syst.}\rm, vol.~16, no.~2, 2019.

\bibitem{pinbot}
S. Noh, \textit{et al.},
{\rm ``Pinbot: A Walking Robot with Locking Pin Arrays for Passive Adaptability to Rough Terrains'',} \textit{Proc. IEEE Int. Conf. Robot. Automat.}, pp.~1438--1444, 2020.

\bibitem{microspine}
A. Asbeck, \textit{et al.},
{\rm ``Scaling Hard Vertical Surfaces with Compliant Microspine Arrays'',} \textit{Int. J. Robot. Res.}\rm, vol.~25, no.~12, pp.~1165--1179, 2006.


\bibitem{uno_jsass}
K. Uno, \textit{et al.},
{\rm ``Qualification of a Time-of-Flight Camera as a Hazard Detection and Avoidance Sensor for a Moon Exploration Microrover'', }\rm \textit{Trans. JSASS, Aerospace Tech. Jpn.}\rm, vol.~16, no.~7, pp.~619--627, 2018.

\bibitem{tanaka_iros2014}
D. Tanaka, \textit{et al.},
{\rm ``Object Manifold Learning with Action Features for Active Tactile Object Recognition'', }\rm \textit{Proc. IEEE/RSJ Int. Conf. Intel. Rob. Syst.}\rm, pp.~608--614, 2014.

\bibitem{fingripper}
J. M. Gandarias, \textit{et al.},
{\rm ``Enhancing Perception with Tactile Object Recognition in Adaptive Grippers for Human–Robot Interaction'', }\rm \textit{Sensors}\rm, vol.~18, no.~3, art.~no.~E692, 2019.

\end{thebibliography}
\end{document}